\documentclass[11pt]{article} 
\usepackage{rldmsubmit,palatino}
\usepackage{graphicx}
\usepackage{amsmath} 
\usepackage{graphicx}
\usepackage{booktabs}
\usepackage{mathtools}
\usepackage{bm}
\usepackage{tabularx}
\usepackage{caption,subcaption}
\usepackage{cite}
\usepackage{siunitx}
\usepackage{xcolor}
\usepackage{float} 
\usepackage{array,multirow}
\usepackage{wrapfig}
\usepackage{hyperref}

\DeclareMathOperator{\sgn}{sgn}

\let\OLDthebibliography\thebibliography
\renewcommand\thebibliography[1]{
  \OLDthebibliography{#1}
  \setlength{\parskip}{1pt}
  \setlength{\itemsep}{0pt}
}

\title{Unlocking the Potential of Simulators: Design with RL in Mind}

\author{
Rika Antonova \quad \quad \quad Silvia Cruciani
\\ \\
KTH -- Royal Institute of Technology\thanks{Robotics, Perception and Learning Group: \texttt{https://www.kth.se/en/csc/forskning/rpl} \quad \quad \quad \quad \quad \quad \quad \quad \quad \quad  \quad \quad  \quad \quad
{\color{white}.}  {\color{white}.} \quad at the School of Computer Science and Communication} \\
Stockholm, Sweden  \\
\texttt{\{antonova, cruciani\}@kth.se} \\
}

\begin{document}

\maketitle



\begin{abstract}
Using Reinforcement Learning (RL) in simulation to construct policies useful in real life is challenging. This is often 
attributed to the \textit{sequential decision making} aspect: inaccuracies in simulation accumulate over multiple steps, hence the simulated trajectories diverge from what would happen in reality. 

In our work we show the need to consider another important aspect: the mismatch in simulating control. We bring attention to the need for modeling control as well as dynamics, since oversimplifying assumptions about applying actions of RL policies could make the policies fail on real-world systems. 

We design a simulator for solving a pivoting task (of interest in Robotics) and demonstrate that even a simple simulator designed with RL in mind outperforms high-fidelity simulators when it comes to learning a policy that is to be deployed on a real robotic system. We show that a phenomenon that is hard to model -- friction -- could be exploited successfully, even when RL is performed using a simulator with a simple dynamics and noise model. Hence, we demonstrate that as long as the main sources of uncertainty are identified, it could be possible to learn policies applicable to real systems even using a simple simulator. 

RL-compatible simulators could open the possibilities for applying a wide range of RL algorithms in various fields. This is important, since currently data sparsity in fields like healthcare and education frequently forces researchers and engineers to only consider sample-efficient RL approaches. Successful simulator-aided RL could increase flexibility of experimenting with RL algorithms and help applying RL policies to real-world settings in fields where data is scarce. We believe that lessons learned in Robotics could help other fields design RL-compatible simulators, so we summarize our experience and conclude with suggestions. 

\end{abstract}


\acknowledgements{
We thank Christian Smith and Danica Kragic for guidance in the ``Pivoting Task'' work, which was supported by the European Union framework program H2020-645403 RobDREAM.}

\startmain 

\section{Introduction}

Using simulators to learn policies applicable to real-world settings is a challenge: important aspects of agent and environment dynamics might be hard and costly to model. Even if sufficient modeling knowledge and computational resources are available, inaccuracies present even in high-fidelity simulators can cause the learned policies to be useless in practice. In the case of \textit{sequential decision making} even a small mismatch between the simulated and the real world could accumulate across multiple steps of executing a policy. 

Despite these challenges, there is potential if simulators are designed with RL in mind. In our work we aim to understand what aspects of simulators are particularly important for learning RL policies. We focus first on the field of Robotics and present an approach to learn robust policies in a simulator, designed to account for uncertainty in the dynamics as well as in control of the real robot. We discuss our simulation and hardware experiments for solving a pivoting task (a robotic dexterity problem of re-orienting an object after initial grasp). Solving this task is of interest to the Robotics community; our approach to solving this task is described in detail in~\cite{antonova2017reinforcement}. Here we only include the aspects of the work relevant to an interdisciplinary RLDM audience and present it as an example of RL-compatible simulator design.

We then discuss the potential of building simulators with RL in mind for other fields. In a number of fields enough domain knowledge is available to construct a task-specific or setting-specific simulator for a particular problem. It would be useful to ensure that such simulators can facilitate learning RL policies that work when transferred to real environments. This would significantly broaden the range of RL algorithms that could be used to solve a given problem -- eliminating the need to restrict attention to only sample-efficient approaches (as is often done, since real-world learning is costly).

\section{Why Use Simulators for Learning?}

Recent success of Reinforcement Learning in games (Atari, Go) and Robotics make RL potentially the next candidate to solve challenging decision making (under uncertainty) problems in a wide variety of fields. A number of attempts to apply RL algorithms to real-world problems have been successful, but the success was mostly limited to sample-efficient approaches. It is costly and time-consuming to search for optimal policies by obtaining real-world trajectories. Batch RL -- a class of algorithms that learn only from previously collected samples -- could be helpful. But batch RL constitutes only a small fraction of a variety of RL algorithms. Hence, to open up the potential for using a wider range of RL algorithms we can turn to simulation. 

A number of fields have already developed general-purpose simulators\footnote{Examples of general-purpose simulators from various fields: SimStudent in education (simstudent.org); NS network simulator used for research in data networks (en.wikipedia.org/wiki/Ns\_(simulator)); more than a dozen general-purpose simulators for robotics (en.wikipedia.org/wiki/Robotics\_simulator).}, and task-specific simulators are frequently developed for solving concrete research problems. But there is a need to facilitate successful transfer of the policies learned in simulation to real-world environments. To achieve this we can either ensure a tight enough match between simulated and real environments and use the learned policies directly, or develop data-efficient ``second-stage'' approaches to adjust the learned policies to the real world.

An additional consideration is current rapid advances in using deep neural networks for RL. Several recently successful algorithms, like Trust Region Policy Optimization (TRPO)~\cite{schulman2015trust} and Deep Deterministic Policy Gradient (DDPG)~\cite{lillicrap2015continuous} are not designed to be particularly sample-efficient. Instead these allow to learn flexible control policies, and can use large neural networks as powerful policy and Q function approximators. In simulation, ``deep RL'' algorithms can solve problems similar in principle to those considered, for example, in Robotics. However, significant practical problems remain before they can be routinely applied to the real-world. Algorithms might not be data-efficient enough to learn on real hardware, so (pre-)training in simulation might be the only feasible solution. Success in simplified simulated settings does not immediately imply the policies could be transferred to a real-world task ``as-is'', but investigating the conditions necessary for successful transfer could enable approaches that succeed in real world environments.

\section{Challenges and Opportunities of Simulation-based Learning: Insights from Robotics}

Robotics presents a great opportunity for research in simulation-to-reality transfer for RL agents. First of all, a wide variety of general and task-specific simulators have been built to simulate various robotic systems. These simulators can model certain aspects well (e.g. the dynamics of rigid bodies with known masses and shapes). However, other aspects are harder to model precisely (e.g. dynamics of \mbox{non-rigid} bodies, friction). Hence, so-called \mbox{``high-fidelity''} simulators contain a combination of models that are precise enough to describe \mbox{real-world} interactions and models that are either imprecise or even misleading. This creates a challenging problem from the perspective of an uninformed RL agent. However, the simplifying aspect is that getting insights into RL algorithms is easier. Experienced roboticists can have a reasonable guess as to which aspect of the environment is hardest for the agent to learn correctly from simulation. This is in contrast with other fields. For example, some leading approaches for modeling student's knowledge in the field of education have error rates not far from random guessing when applied to non-synthetic student data. So the modeling problem is hard, but it is not at all clear which aspect is at fault and what exactly is making the real student data so challenging. Another example: modeling traffic in data networks requires significant simplifications for large-scale (internet-scale) simulators. This causes difficulties if an RL agent is learning a policy for handling individual packets. By comparison, robotic simulators provide a more clear picture of what would or would not be modeled well, and there is a way to quickly test policies learned in simulation on the real environment. Such robotics experiments are frequently more accessible than running large-scale studies involving humans, where returns from the policy could be distant in time (e.g. learning outcome of a class lasting several months) or too high-stakes for experimentation (result of a treatment for a patient). Thus one strategy is to develop and test a variety of simulation-to-reality transfer  approaches on robotic tasks, then use the promising approaches for costlier and longer-term studies in other fields. \\

We note that care should be taken to capture (even if only approximately) all the relevant environment and control dynamics in a simulator. Roboticists have looked into problems arising from under-modeling of dynamics and hardware wear-and-tear (e.g.~\cite{atkeson1994using}) and experimented with various approaches of injecting noise in the deterministic simulators to increase tolerance to slight modeling inaccuracies (~\cite{kober2013reinforcement} gives a summary and further references). Nonetheless, one aspect that is frequently overlooked is the application of control actions to real systems. Simple systems built in the lab can be designed to support fast and direct control. However, as the field moves to more sophisticated robotic systems that are built for a variety of tasks -- oversimplifying assumptions of control could render simulation useless. Consider, for example, the Baxter robot -- a medium-to-low-cost system built for small businesses and research labs. This robot is built to be relatively inexpensive, safe and versatile. Hence, in its usual operating mode, the control of the robot differs from that of high-cost industrial robots, and from that of simpler custom-built research systems. Speed and precision of controlling Baxter is traded off for lower cost of components, safe and smooth motions, versatility.

In our work, in addition to incorporating previously proposed solutions to deal with policy transfer challenges, we also emphasize the need to model delays and inaccuracies in control. We envision that as more multi-purpose medium-to-low-cost robots become widely used, it should become a standard component in building simulators. With that, we demonstrate that even a simple simulator designed with this in mind can be suitable for simulation-based RL.

\section{Building RL-compatible Simulator for Solving a Pivoting Task}

In this section we present a short summary of our research work described in~\cite{antonova2017reinforcement}, with an intent to give an example that is comprehensible to an interdisciplinary audience. We describe our approach to building a task-specific simulator for a pivoting task, then learning an RL policy in simulation and using it on the real robot\footnote{Video summarizing hardware experiments is available at \texttt{https://www.youtube.com/watch?v=LWSjYI9a9xw}} (without further adjustments).

\begin{wrapfigure}{l}{0.3\textwidth}
\vspace{-10px}
\includegraphics[height=0.2\textwidth]{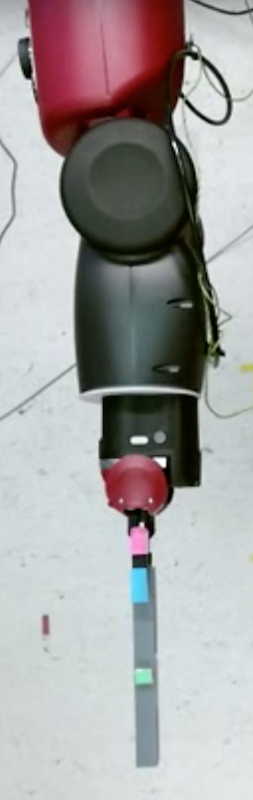}
\hspace{5px}
\includegraphics[height=0.2\textwidth]{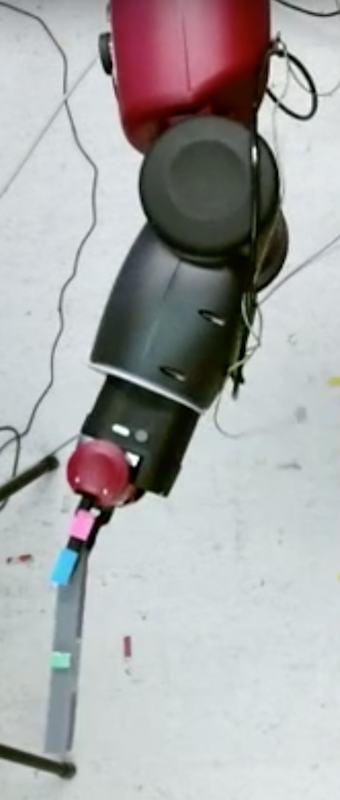}
\hspace{5px}
\includegraphics[height=0.2\textwidth]{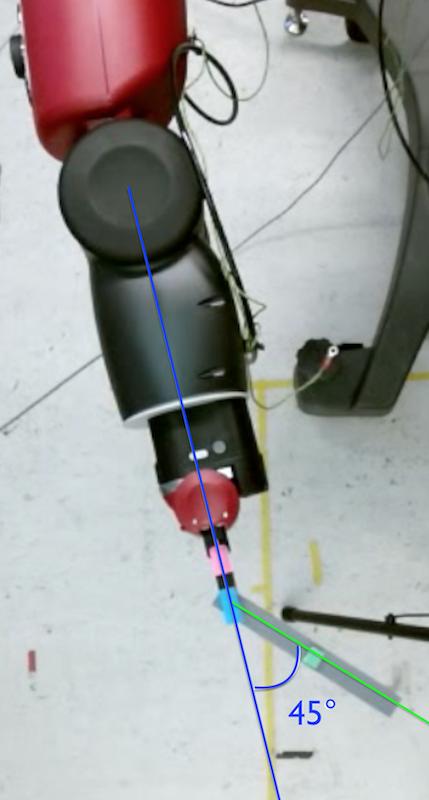}
\caption{\small Pivoting task on Baxter robot.}\label{fig_baxter}
\vspace{-15px}
\end{wrapfigure}

The objective of a pivoting task is to pivot a tool to a desired angle while holding it in the gripper. This can be done by moving the arm of the robot to generate inertial forces sufficient to move the tool and, at the same time, opening or closing the gripper's fingers to change the friction at the pivoting point (gaining more precise control of the motion). The robot can use a standard planning approach to grasp the tool at a random angle $\phi_{init}$. The goal is to pivot it to a target angle $\phi_{tgt}$. 

The limitations of the previous approaches include: only open loop control with no control of the gripping force~\cite{holladay}, movement restricted to the vertical plane~\cite{sintov}, gripper in a fixed position, thus motion of the tool is determined only by the gravitational torque and torsional friction~\cite{vina_2}. All these prior approaches rely strongly on having an accurate model of the tool, as well as precise measurement and modeling of the friction. Since it is difficult to obtain these, we develop an alternative.

\subsection{Dynamic and Friction Models}
\label{subsec_model_simulator}

\begin{wrapfigure}{l}{0.23\textwidth}
\vspace{-10px}
\includegraphics[width=0.23\textwidth]{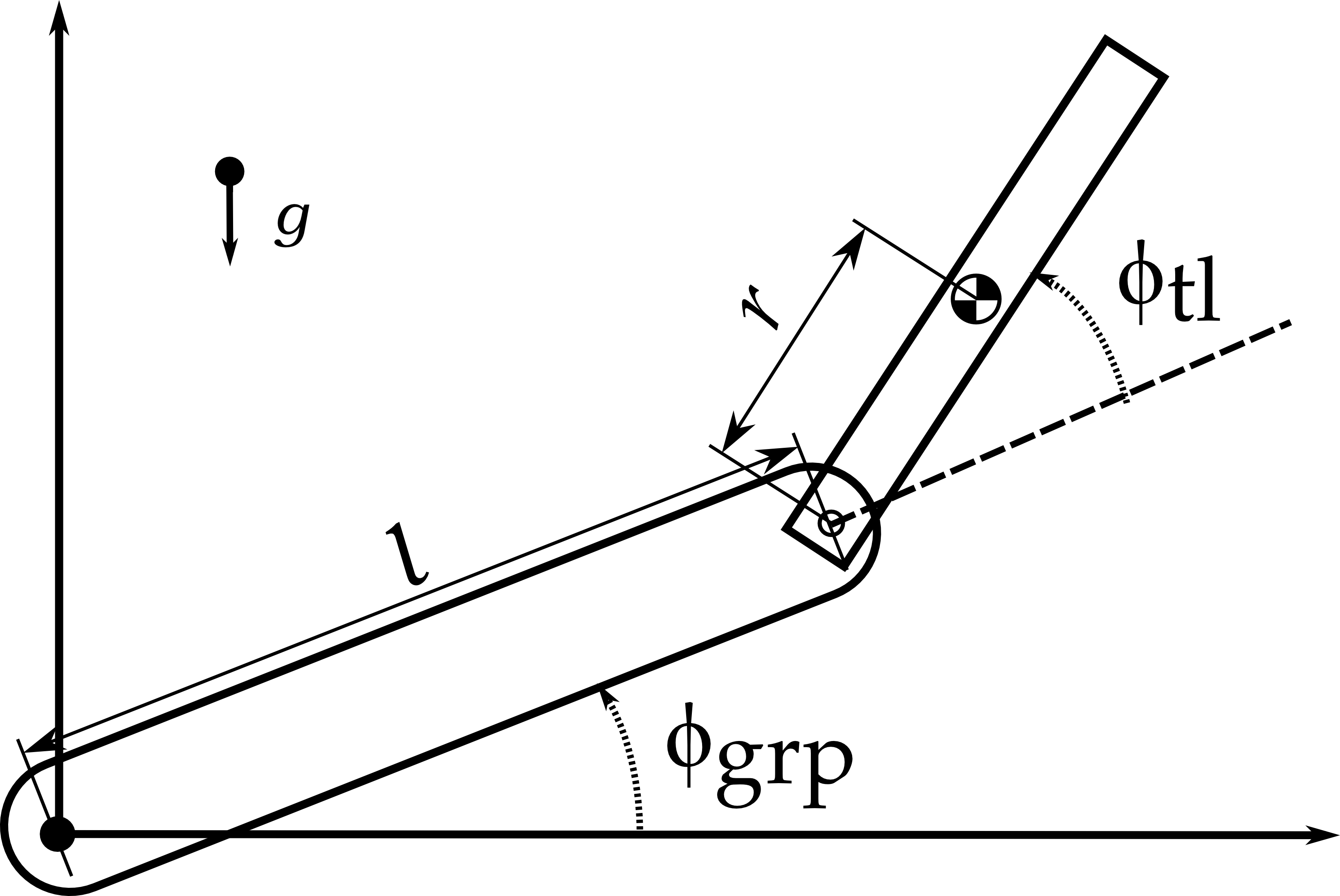}
\caption{\small Model of a 2-link planar arm. First link is the gripper, second is the tool rotating around pivoting point.}\label{fig_model}
\vspace{-20px}
\end{wrapfigure}

Our simulator system is composed of a parallel gripper attached to a link that can rotate around a single axis. This system is an under-actuated two-link planar arm, in which the under-actuated joint corresponds to the pivoting point.  
We assume that we can control the desired acceleration on the first joint.
The dynamic model of the system is given by:
\begin{equation} \label{dynamic_model}
\begin{split}
(I\text{+}mr^2\text{+}mlr\cos(\phi_{tl}))\ddot{\phi}_{grp}+(I\text{+}mr^2)\ddot{\phi}_{tl}+mlr\sin(\phi_{tl})\dot{\phi}_{grp}^2+mgr\cos(\phi_{grp}\text{+}\phi_{tl}) = \tau_f,
\end{split}
\end{equation}
where the variables are as follows: 
$\phi_{grp}$ and $\phi_{tl}$ are the angles of the first and second link respectively; $\ddot{\phi}_{grp}$ and $\ddot{\phi}_{tl}$ are their angular acceleration and $\dot{\phi}_{grp}$ is the angular velocity of the first link; $l$ is the length of the first link; $I$ is the tool's moment of inertia with respect to its center of mass; $m$ is the tool's mass; $r$ is the distance of its center of mass from the pivoting point; $g$ is the gravity acceleration; $\tau_f$ is the torsional friction at the contact point between the gripper's fingers and the tool. 
The second link represents the tool and $\phi_{tl}$ is the variable we aim to control. Figure~\ref{fig_model} illustrates this model.

Pivoting exploits friction at the contact point between the gripper and the tool to control the rotational motion. Such friction is controlled by enlarging or tightening the grasp.  
When the tool is not moving ($\dot{\phi}_{tl}\!\!=\!\!0$), the static friction $\tau_s$ is modeled according to the Coulomb friction model: $|\tau_s|\leq\gamma f_n$, where $\gamma$ is the coefficient of static friction and $f_n$ is the normal force applied by gripper's fingers on the tool.
When the tool moves with respect to the gripper, we model the friction torque $\tau_f$ as viscous friction and Coulomb friction \cite{olsson}: $\tau_f\!=\!-\!\mu_v\dot{\phi}_{tl}\!-\!\mu_cf_n\sgn(\dot{\phi}_{tl})$,
in which $\mu_v$ and $\mu_c$ are the viscous and Coulomb friction coefficients and $\sgn(\cdot)$ is the signum function.
Since most robots are not equipped with tactile sensors to measure the normal force $f_n$ at the contact point, as in~\cite{vina_1} we express it as a function of the distance $d_{fing}$ between the fingers using a linear deformation model: $f_n\!=\!k(d_0\!-\!d_{fing})$,
where $k$ is a stiffness parameter, $d_0$ is the distance at which fingers initiate contact with the tool. 

\subsection{Learning Robust Policies}

We formulate the pivoting task as an MDP. The state space is comprised of states observed at each time step $t$: $s_t\!=[ \phi_{tl} - \phi_{tgt} \ ,\ \dot{\phi}_{tl} \ ,\  \phi_{grp} \ ,\  \dot{\phi}_{grp} \ ,\  d_{fing} ]$, with notation as in the previous subsection.
The actions are: $a_t=\{\ddot{\phi}_{grp}, d_{fing} \}$, where $\ddot{\phi}_{grp}$ is the rotational acceleration of the robot arm, and $d_{fing}$ is the direction of the desired change in distance between the fingers of the gripper. 
The state transition dynamics is implemented by the simulator, but the RL algorithm does not have an explicit access to these.
Reward is given at each time step $t$ such that higher rewards are given when the angle of the tool is closer to the target angle: $r_t = \frac{- |\phi_{tl} - \phi_{tgt}|}{\phi_{RNG}} \in [-1,1]$ ($\phi_{RNG}$ is a normalizing constant). A bonus reward of 1 is given when the tool is close to the target and stopped.

We aim to learn from simulated environment, while being robust to the discrepancies between the simulation and  execution on the robot. For this purpose, we first built a simple custom simulator using equations from Section~\ref{subsec_model_simulator}. To facilitate learning policies robust to uncertainty, we added 10\% noise to friction values estimated for the tool modeled by the simulator. We also injected up to 10\% randomized delay for arm and finger actions in simulation: used noisy linear increase in velocity (as response to acceleration action) and simulated changing fingers' distance with noisy steps. 

We then trained a model-free deep RL policy search algorithm TRPO~\cite{schulman2015trust} on our simulated setting. TRPO has been shown to be competitive with (and sometimes outperform) other recent continuous state and action RL algorithms~\cite{duan2016benchmarking}. However, to our knowledge it has not yet been widely applied to real-world robotics tasks. While the background for the algorithm is well-motivated theoretically, the approximations made for practicality, along with challenges in achieving reasonable training results with a small-to-medium number of samples, could impair the applicability of the algorithm to learning on the robot directly. Hence we explore the approach of using TRPO for policy search in a simulated environment. 

Our model of delayed and noisy control matched the behavior of real Baxter better than assuming near-instantaneous high-precision control. In contrast, learning from a high-fidelity simulator (V-REP) yielded policies that failed completely on the real robot: the tool was swayed in random directions, dropped. V-REP's oversimplification of simulating control impaired the ability of RL to learn useful policies from what was considered to be a relatively high-fidelity simulation.

\subsection{Summary of Experiments}

\begin{wrapfigure}{r}{0.34\textwidth}
\centering
\vspace{-10px}
\caption{\small{Evaluation of TRPO training. Initial and target angles chosen randomly from $[\frac{-\pi}{2},\frac{\pi}{2}]$, training iterations had 50 episodes.}}
\label{fig_train_progress}
\vspace{-7px}
\includegraphics[width=0.31\textwidth]{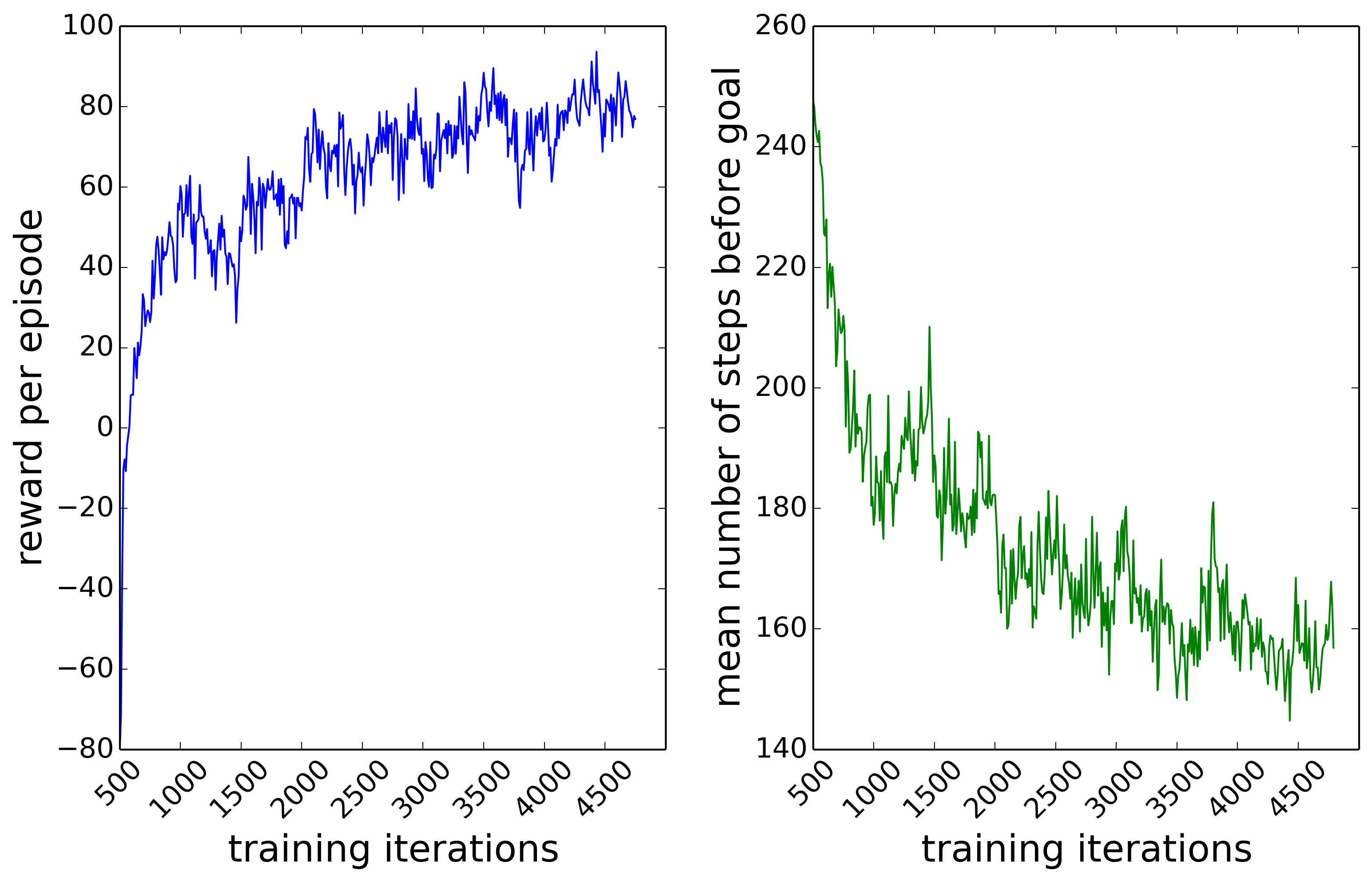}
\vspace{-20px}
\end{wrapfigure}

We trained TRPO (using rllab~\cite{duan2016benchmarking} implementation as a starting point and parameters as reported in~\cite{schulman2015trust} for simulated control tasks) with a fully connected network with 2 hidden layers (with 32, 16 nodes) for policy and Q function approximators. 
The motion was constrained to be linear in a horizontal plane. However, since the learning was not at all informed of the physics of the task, any other plane could be chosen and implemented by the simulator if needed. 

To simulate the dynamics of the gripper arm and the tool, we used the modeling approach from Section~\ref{subsec_model_simulator} and injected noise during training to help learn policies robust to mismatch between the simulator and the robot. While we allowed only up to 10\% noise in the variable modeling Coulomb friction, we observed that the policy was still able to retain 40\% success rate of reaching the target angle when tested on settings with 250\% to 500\% change. This level of robustness alleviates the need to estimate friction coefficients precisely. 

\begin{wrapfigure}{r}{0.34\textwidth}
\centering
\caption{\small{Distance to target vs time for experiments on Baxter averaged over all trials (30 trials for each tool).}}
\label{baxter_experiments}
\vspace{-7px}
\includegraphics[width=0.32\textwidth]{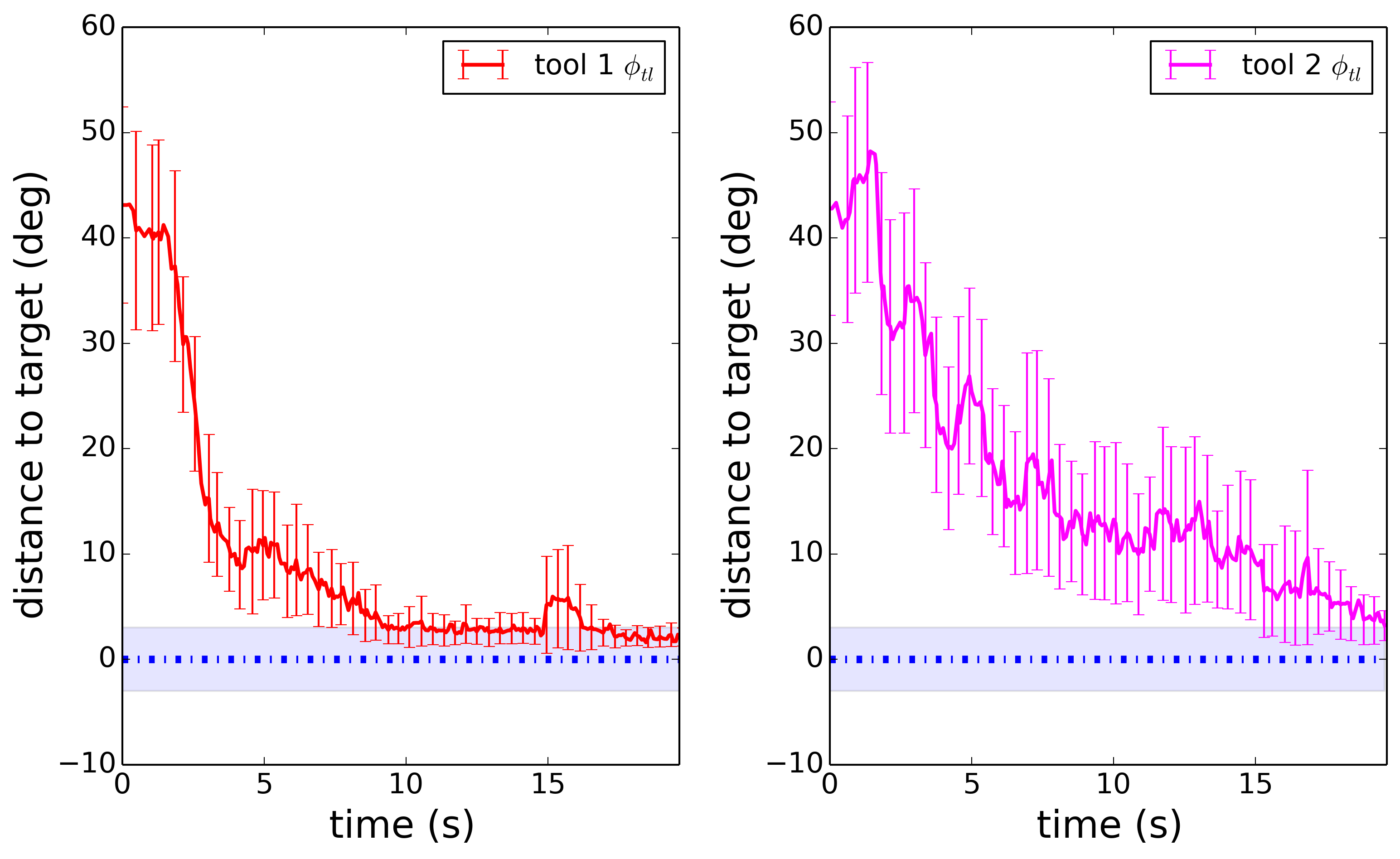}
\vspace{-12px}
\end{wrapfigure}

We deployed the learned policies on a Baxter robot and ran experiments with two objects of different materials, hence different friction properties. The friction coefficients used for modeling the tool in the simulator have been estimated only for the first tool. Hence the policy was not explicitly trained using the parameters matching those of the second tool.

We obtained a $\approx\!\!93\%$ success rate with tool~1 and $\approx\!\!83\%$ with tool~2. As expected, the policy performs better with the tool whose parameters were used (in a noisy manner) during training in simulation. Nonetheless, the performance using the tool not seen during training was robust as well. 
Fig.~\ref{baxter_experiments} illustrates the performance averaged over all the trials. 
The target is reached faster when tool~1 is used, and a bit slower when tool~2 is used.
The deviation from the goal region after reaching the goal is likely due to inaccuracies in visual tracking. After reporting that the tool is in the goal region, the tracker might later report a corrected estimate, indicating further tool adjustment is needed. We observe that in such cases the policy still succeeds in further pivoting the tool to the target angle.

\section{Conclusion \& Implications for other Fields for Designing RL-compatible Simulators}

Our experience of building an RL-compatible simulator for a particular robotics task yields a few general suggestions. 

\textit{1. Focused modeling of the dynamics}: Since high-fidelity modeling could be expensive, one could consider modeling only the part of the dynamics relevant to the task. Domain expertise could be used to understand what is ``relevant''. E.g. high-fidelity modeling of the robot arm dynamics might be unnecessary, if the task could be solved by fixing the motion to a particular plane where even simple dynamics equations are sufficient. In fields that already have general-purpose simulators such ``focusing'' is not always easy. The design usually does not allow to make a general-purpose simulator more task-oriented. Even if an researcher can find a restriction to the case where dynamics should be simpler -- there is frequently no option in the simulator to eliminate unnecessary computations. Hence, the suggestion is to keep in mind this design consideration when building general-purpose simulators: either give the users easy control of which simulation modules to turn on/off or run automatic internal optimization of the simulator.

\textit{2. Not oversimplifying the model of executing actions}: Neglecting to observe that the process of applying an action needs modeling could cause policies learned in even a high-fidelity simulator to fail on a real system. In contrast, including even a simple model of control application in the simulator could fix the problem. Another solution could be to re-formulate an MDP/POMDP for the task to make primitive actions be something that is trivial to apply to the real system (and capture all the side-effects of applying a primitive action in the dynamics of the state transitions). This solution could work, though it might require unintuitive formulations. Regardless of the approach, the main insight is not to neglect this aspect altogether (control of real robots is not instantaneous; the action of "give medication A" might not lead to state ``took medication A'' on the patient's side; the action of ``show advertisement B'' might be blocked/ignored by the users).

\textit{2. Even simple noise models help if sources of variability/uncertainty are identified}: Adding Gaussian noise to various parts of a deterministic dynamical system is an approach frequently used in robotics to ``model'' uncertainty. No doubt this could be too simplistic for some settings/tasks. However, since learning in simulation opens potential for cheaper experimentation -- it is possible to try various ways of capturing uncertainty and identify those sufficient for learning successful policies. As we show in our experiments, a phenomenon that is hard to model -- friction -- could be still exploited successfully for control even with a simple dynamical model and a simple noise model. As long as the main sources of uncertainty are identified, it could be possible to learn policies applicable to real systems even from a simple simulator.

{\tiny
\bibliographystyle{plain}
\bibliography{Pivoting,DexterousManip,FrictionControl,RLinRobotics,General}

\begin{thebibliography}{10}

\bibitem{antonova2017reinforcement}
Rika Antonova, Silvia Cruciani, Christian Smith, and Danica Kragic.
\newblock Reinforcement learning for pivoting task.
\newblock {\em arXiv preprint arXiv:1703.00472}, 2017.

\bibitem{atkeson1994using}
Christopher~G Atkeson et~al.
\newblock Using local trajectory optimizers to speed up global optimization in
  dynamic programming.
\newblock {\em Advances in neural information processing systems}, pages
  663--663, 1994.

\bibitem{duan2016benchmarking}
Yan Duan, Xi~Chen, Rein Houthooft, John Schulman, and Pieter Abbeel.
\newblock Benchmarking deep reinforcement learning for continuous control.
\newblock {\em arXiv preprint arXiv:1604.06778}, 2016.

\bibitem{holladay}
A.~Holladay, R.~Paolini, and M.~T. Mason.
\newblock A general framework for open-loop pivoting.
\newblock In {\em 2015 IEEE International Conference on Robotics and Automation
  (ICRA)}, pages 3675--3681, May 2015.

\bibitem{kober2013reinforcement}
Jens Kober, J~Andrew Bagnell, and Jan Peters.
\newblock Reinforcement learning in robotics: A survey.
\newblock {\em The International Journal of Robotics Research}, page
  0278364913495721, 2013.

\bibitem{lillicrap2015continuous}
Timothy~P Lillicrap, Jonathan~J Hunt, Alexander Pritzel, Nicolas Heess, Tom
  Erez, Yuval Tassa, David Silver, and Daan Wierstra.
\newblock Continuous control with deep reinforcement learning.
\newblock {\em arXiv preprint arXiv:1509.02971}, 2015.

\bibitem{olsson}
H.~Olsson, K.~J. Åström, M.~Gäfvert, C.~Canudas~De Wit, and P.~Lischinsky.
\newblock Friction models and friction compensation.
\newblock {\em Eur. J. Control}, page 176, 1998.

\bibitem{schulman2015trust}
John Schulman, Sergey Levine, Philipp Moritz, Michael~I Jordan, and Pieter
  Abbeel.
\newblock Trust region policy optimization.
\newblock {\em CoRR, abs/1502.05477}, 2015.

\bibitem{sintov}
A.~Sintov and A.~Shapiro.
\newblock Swing-up regrasping algorithm using energy control.
\newblock In {\em 2016 IEEE International Conference on Robotics and Automation
  (ICRA)}, pages 4888--4893, May 2016.

\bibitem{vina_1}
F.~E. Vi{\~{n}}a, Y.~Karayiannidis, K.~Pauwels, C.~Smith, and D.~Kragic.
\newblock In-hand manipulation using gravity and controlled slip.
\newblock In {\em Intelligent Robots and Systems (IROS), 2015 IEEE/RSJ
  International Conference on}, pages 5636--5641, Sept 2015.

\bibitem{vina_2}
F.~E. Vi{\~{n}}a, Y.~Karayiannidis, C.~Smith, and D.~Kragic.
\newblock Adaptive control for pivoting with visual and tactile feedback.
\newblock In {\em 2016 IEEE International Conference on Robotics and Automation
  (ICRA)}, pages 399--406, May 2016.

\end{thebibliography}
} 

\end{document}